\documentclass[10pt,twocolumn,letterpaper]{article}

\usepackage[pagenumbers]{cvpr} % To force page numbers, e.g. for an arXiv version

\definecolor{cvprblue}{rgb}{0.21,0.49,0.74}
\usepackage[pagebackref,breaklinks,colorlinks,allcolors=cvprblue]{hyperref}

\def\paperID{****} % *** Enter the Paper ID here
\def\confName{CVPR}
\def\confYear{2026}

\title{TalkVerse: Democratizing Minute-Long Audio-Driven Video Generation}

\author{Zhenzhi Wang$^{1}$, Jian Wang$^2$, Ke Ma$^2$, Dahua Lin$^1$, Bing Zhou$^2$ \\
$^1$The Chinese University of Hong Kong, $^2$Snap Inc.   \\
 \small\texttt{wz122@ie.cuhk.edu.hk}
}
\begin{document}
\maketitle
\begin{abstract}
We introduce TalkVerse, a large-scale, open corpus for single-person, audio-driven talking video generation designed to enable fair, reproducible comparison across methods. While current state-of-the-art systems rely on closed data or compute-heavy models, TalkVerse offers 2.3 million high-resolution (720p/1080p) audio-video synchronized clips totaling 6.3k hours. These are curated from over 60k hours of video via a transparent pipeline that includes scene-cut detection, aesthetic assessment, strict audio-visual synchronization checks, and comprehensive annotations including 2D skeletons and structured visual/audio-style captions. Leveraging TalkVerse, we present a reproducible 5B DiT baseline built on Wan2.2-5B. By utilizing a video VAE with a high downsampling ratio and a sliding window mechanism with motion-frame context, our model achieves minute-long generation with low drift. It delivers comparable lip-sync and visual quality to the 14B Wan-S2V model but with 10$\times$ lower inference cost. To enhance storytelling in long videos, we integrate an MLLM director to rewrite prompts based on audio and visual cues. Furthermore, our model supports zero-shot video dubbing via controlled latent noise injection. We open-source the dataset, training recipes, and 5B checkpoints to lower barriers for research in audio-driven human video generation. Project Page: \url{https://zhenzhiwang.github.io/talkverse/}.
\end{abstract} 
\section{Introduction}
\begin{figure*}[t]
    \centering
    \includegraphics[width=\textwidth]{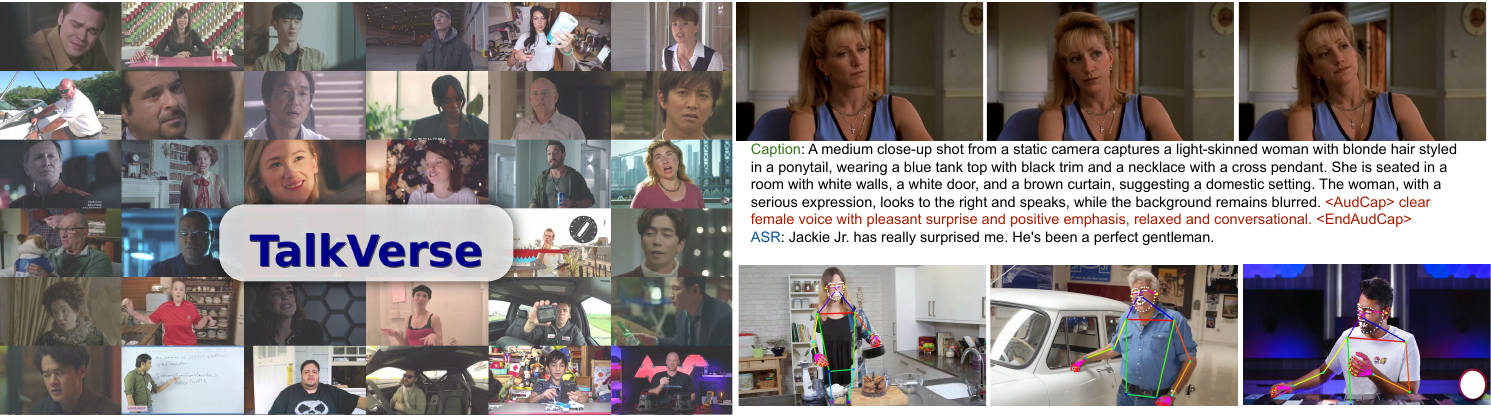}
        \vspace{-2em}
    \caption{Overview of TalkVerse with 2.3M single-person clips (6.3K hours) at 720p with synchronized speech audio, 2D skeleton sequences, and structured captions (visual+audio-style). Dataset and training recipes for the 5B baseline for audio-driven video generation will be open-sourced. It can also facilitate audio-video joint generation, pose-driven video generation, and video personalization. (top right) Video examples paired with visual+audio captions, as well as ASR results. (bottom right) Pose visualizations.}
    \label{fig:teaser}
    \vspace{-1em}
    \end{figure*}
Human video generation from audio and pose conditions~\citep{he2023gaia,emo,xu2024hallo,wang2024vexpress,chen2024echomimic,xu2024vasa,stypulkowski2024diffused,jiang2024loopy,lin2024cyberhost,lin2025omnihuman,ding2025kling,chen2025hunyuanvideo} has advanced rapidly with pretrained video diffusion models~\citep{bar2024lumiere,blattmann2023svd,ayl,guo2024animatediff,zhou2022magicvideo,walt,wang2023modelscope,ho2022video,videogan,cvideogan,singer2022make,text2video,villegas2022phenaki,lin2025apt}, enabling expressive talking/performing portraits and beyond. Yet high-quality research and applications in this space remain hampered by (1) limited open training data with reliable audio-visual synchronization, and (2) compute-heavy pretrained models (e.g., Wan2.1-14B) that impede broad participation. In this work, we show that curating a large-scale and transparent full-body human-centric video corpus with high-quality synchronized audio is feasible, and it could foster the training of lightweight models with lip synchronization and visual quality on par with commercial-level models, by leveraging limited computational resources (e.g., 64 GPUs for one week) with the goal of reproducibility at moderate compute, not SOTA-at-any-cost.

Although many existing methods managed to generate full-body audio-driven videos~\cite{lin2025omnihuman,ding2025kling,chen2025hunyuanvideo,tu2025stableavatar,gan2025omniavatar,gao2025wan}, their training datasets are either internal data~\cite{lin2025omnihuman,ding2025kling,jiang2025omnihuman,chen2025hunyuanvideo} or a combination of open-source talking-head datasets~\cite{zhu2022celebv,DBLP:journals/tog/EphratMLDWHFR18,hallo3,hdtf} and a limited amount of self-collected full-body data. Besides, the recent large-scale human-centric dataset OpenHumanVid~\cite{li2024openhumanvid} focuses on text-to-video generation and ignores the audio synchronization property for human image animation. We identify several challenges with existing open-source data: \textbf{(1) Video-audio synchronization}: Unlike pose and text conditions that can be inferred from video by pose detection or video captioning, the audio modality is recorded during collection and synchronization is not guaranteed in internet videos; \textbf{(2) Appearance consistency and disambiguation}: it is hard to train audio-driven video generation models with ambiguity of the active speaker when multiple subjects exist; \textbf{(3) Scale of data}: as we want to generate realistic lip and full-body movements from in-the-wild audio, the total duration of data and language coverage are vital for this task. The assumption is, the training corpus should cover most phonemes across many languages so that the model can generalize. Therefore, a high-quality and large-scale dataset is necessary to facilitate this area and end-to-end reproducibility.

In this paper, we introduce TalkVerse, a large-scale audio-video corpus for single-person, full-body human video generation conditioned on audio or pose. It comprises \emph{2.3M} high-resolution (720p or 1080p) audio-synchronized clips with over \emph{6.3k} hours from two human-centric text-to-video sources~\cite{li2024openhumanvid,chen2024panda} with 30M initial human-related clips, as shown in Fig.~\ref{fig:teaser}. It is curated through a pipeline that performs scene-cut detection, human and subtitle screening, aesthetic and image quality assessment, motion filtering, and audio-video synchronization checks, followed by language annotation. It emphasizes verified audio-video synchronization and high-quality, identity- and motion-diverse content. To preserve most of the text-conditioning ability when training audio-conditioning injection from a pretrained Wan DiT~\cite{wang2025wan}, we also annotate structured video captions with background appearance, camera movement, subject appearance, and subject movement from a fine-tuned Qwen2.5-VL~\cite{bai2025qwen2} and an audio-style caption from Qwen3-Omni~\cite{xu2025qwen3}. We utilize a pose detector~\cite{dwpose} to ensure there is a single subject and also extract the pose condition. In summary, we provide various human-related annotations, including video captions, synchronized audio, 2D human pose, audio-style captions, and language type. Given the properties of audio-video synchronization, this dataset is also valuable for other human-centric video tasks, such as audio-video joint generation, pose-driven video generation, and video personalization from human images.

To demonstrate the utility of the data, we adopt the Wan-S2V architecture~\citep{gao2025wan} and train a 5B model using a higher VAE downsampling ratio. Despite being over 10$\times$ cheaper at inference than the strong Wan-S2V-14B, this baseline attains comparable perceptual quality and lip synchronization. In summary, our contributions are: \textbf{(1) TalkVerse-2.3M.} We release a large, synchronized, single-person talking/performing video corpus (with \emph{2.3M} clips and \emph{6.3k}+ hours) curated by a transparent pipeline with reproducible filters for visual quality, motion, and audio-video alignment. \textbf{(2) TalkVerse-5B model and open recipes.} We open-source data recipes, preprocessing tools, training scripts, and inference code, along with checkpoints for a compact 5B Wan-S2V baseline trained end-to-end on TalkVerse.

\section{Related Work}
\noindent {\bf Human Animation Models} synthesize human videos conditioned on text, reference images, human body poses, or audio. Early GAN-based approaches~\citep{siarohin2019fomm, zhao2022tps, mraa, wang2021facev2v}, typically trained on comparatively small datasets~\citep{nagrani2017voxceleb, siarohin2019fomm, xie2022vfhq, zhu2022celebv}, tackled self-supervised pose transfer. Recent diffusion-based methods~\citep{shao2024human4dit, zhang2024mimicmotion, aa,DBLP:conf/nips/00010ZF0LTCX0L24,wang2025multi} surpass GANs by conditioning on 2D skeletons, estimated depth, or mesh sequences. Audio-driven portrait animation~\citep{geneface, zhang2023sadtalker, emo, jiang2024loopy, hallo3} has expanded to full-body motion through both two-stage pipelines~\citep{vlogger, DBLP:conf/cvpr/MengZLM25, emo2, diffted} or unified one-stage designs~\citep{lin2024cyberhost,lin2025omnihuman, tu2025stableavatar,gan2025omniavatar, chen2025hunyuanvideo,ding2025kling, wang2025interacthuman,gao2025wan}. However, many prior works remain for internal use or are released only as checkpoints. In contrast, we are the first to disclose the full data-curation pipeline for training high-quality human animation models. Our dataset constitutes, to our knowledge, the first million-scale open-source resource for this task.

\begin{figure*}[t]
    \centering
    \includegraphics[width=0.9\textwidth]{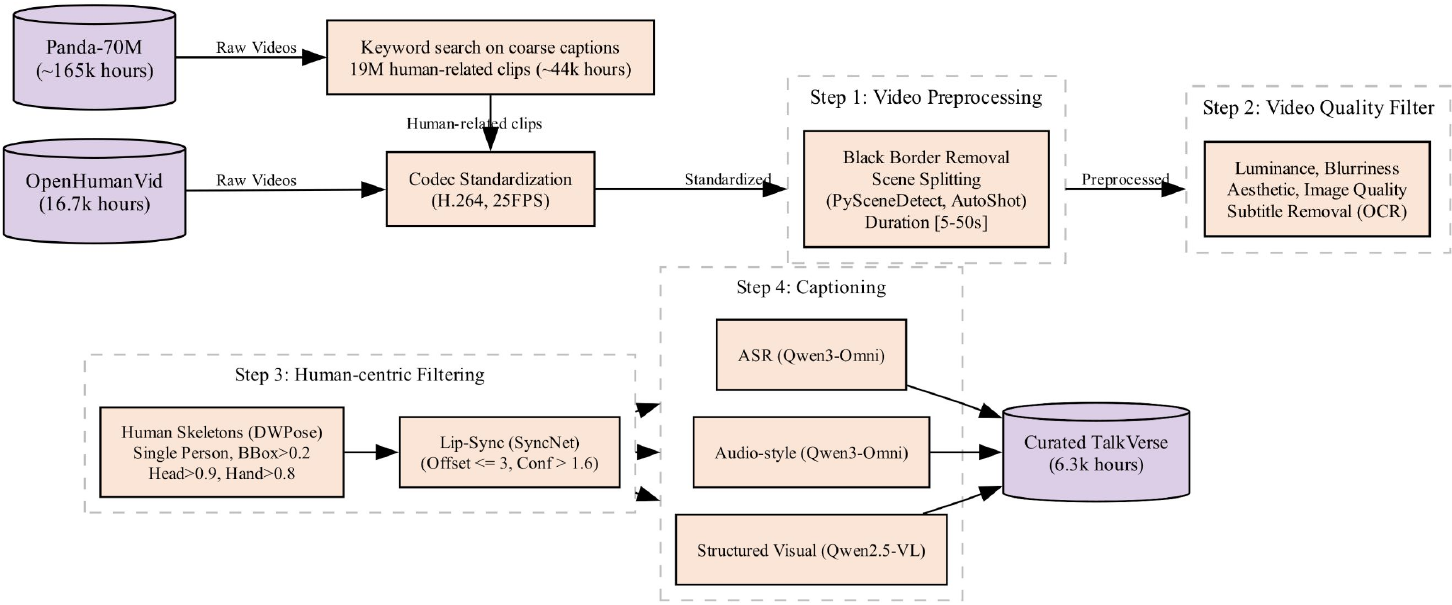}
        \vspace{-1em}
    \caption{Overview of our comprehensive data filtering pipeline. We obtain 6.3k hours of high-quality audio--video data from 60k hours of human-related videos, and provide 2D skeletons and extensive text annotations from three aspects: visual, audio-style, and ASR content.}
    \label{fig:data_pipeline}
    \vspace{-1em}
    \end{figure*}

\noindent {\bf Human-Centric Video Datasets} with paired, temporally synchronized modalities (e.g., pose, audio, and text) are critical for progress in human image animation. Prior human-centric datasets either prioritize pose without audio~\cite{jafarian2021learning, zablotskaia2019dwnet,DBLP:conf/nips/00010ZF0LTCX0L24,DBLP:conf/cvpr/BlackPTY23} or are limited to talking heads~\cite{zhu2022celebv,hallo3,hdtf}. Meanwhile, action-recognition datasets such as UCF101~\cite{soomro2012ucf101}, NTU RGB+D~\cite{shahroudy2016ntu}, ActivityNet~\cite{caba2015activitynet}, and AVA~\cite{gu2018ava} suffer from low resolution and reduced visual quality for our purposes. General-purpose text-to-video corpora, e.g., HD-VILA-100M~\cite{xue2022advancing}, HowTo100M~\cite{miech2019howto100m}, WebVid-10M~\cite{bain2021frozen}, Panda-70M~\cite{chen2024panda}, and OpenVid-1M~\cite{nan2024openvid}, lack diversity within the human category (identities and motions) and typically do not provide accurate captions, skeleton representations, or synchronized speech audio as motion conditions. More recently, OpenHumanVid~\cite{li2024openhumanvid} introduced 16.7k hours of human-centric video, but its duration and visual quality are still limited and audio is not guaranteed to be synchronized for animation training. TalkCuts~\cite{chen2025talkcuts} collects multi-shot, long-form talk shows, yet the scale remains about 500 hours. While valuable, these resources are insufficient for training human image animation models due to constraints in diversity, scale, audio synchronization, and skeleton quality. Building on general text-to-video corpora such as Panda-70M and OpenHumanVid, we curate a high-quality, single-person full-body human-centric dataset with verified pose quality and audio-pose synchronization, and demonstrate its effectiveness for training human image animation models in our experiments.
 
\begin{table*}[t]
    \centering
    \caption{Comparison of representative datasets for human-centric video understanding and generation. 
    We integrate skeleton sequences derived from DWPose~\cite{dwpose} and ensure the speech audio is synchronized through SyncNet~\cite{syncnet}. In audio-driven video generation, we are the largest dataset with synchronized speech audio and keep full-body human movements instead of only talking heads.}
    \vspace{-0.5em}
    \resizebox{0.9\textwidth}{!}{
    \begin{tabular}{@{}cccccc@{}}
    \toprule
     Dataset name & Domain & \# Videos & Length (hours) & Motion type & Resolution \\ 
    \midrule
      WebVid-10M\cite{bain2021frozen} & Open & 10M & 52K & Text & 360P \\
      Panda-70M\cite{chen2024panda} & Open & 70M & 167K & Text & 720P \\
      OpenVid-1M\cite{nan2024openvid} & Open & 1M & 2K & Text & 512P \\
      \midrule
      UCF-101\cite{soomro2012ucf101} & Human action & 13.3K & 26.7 & Text & 240P \\
      NTU RGB+D\cite{shahroudy2016ntu} & Human action & 114K & 37 & 3D pose, depth & 1080P \\
      ActivityNet\cite{caba2015activitynet} & Human action & 100K & 849 & Text & 480P \\
      \midrule
      TikTok\cite{DBLP:conf/cvpr/JafarianP21} & Single-person dance & 350 & 1 & Skeleton & 600P \\
      UBC-Fashion\cite{zablotskaia2019dwnet} & Single-person dance & 500 & 2 & Skeleton & 720P \\
      HumanVid\cite{DBLP:conf/nips/00010ZF0LTCX0L24} & Single-person dance & 20K & 115 & Skeleton, camera & 1080P \\
    OpenHumanVid\cite{li2024openhumanvid} & Multi-person performing & 13.2M&16.7K & Text, skeleton pose, non-sync audio & 720P \\
      \midrule
      VoxCeleb2\cite{DBLP:conf/interspeech/ChungNZ18} & Talking heads & 1M & 2.4K & speech audio & 240P \\
      HDTF\cite{hdtf} & Talking heads & 368 & 15.8 & speech audio & 512P \\
      TalkingHead-1KH\cite{wang2021facev2v} & Talking heads & 80K & 160 & speech audio & 512P \\
      CelebV-HQ\cite{zhu2022celebv} & Talking heads & 35K & 68 & Text, speech audio & 512P \\
      CelebV-Text\cite{yu2023celebv} & Talking heads & 70K & 279 & Text, speech audio & 512P \\
      DH-FaceVid-1K\cite{di2025dh} & Talking heads & 270K & 1.2K & Text, speech audio & 512P \\
      TalkCuts~\cite{chen2025talkcuts} &  Single-person talking & 164k & 507 & Text, skeleton pose, speech audio & 1080P \\
      \midrule
      Ours & Single-person talking & 2.3M & 6.3K & Text, skeleton pose, speech audio & 720P/1080P \\
    \bottomrule
    \end{tabular}
    }
    \vspace{-1em}
    \label{tab:datasets}
    \end{table*}

\begin{figure*}[t]
    \centering
    \begin{subfigure}{0.42\linewidth}
        \centering
        \includegraphics[width=0.99\textwidth]{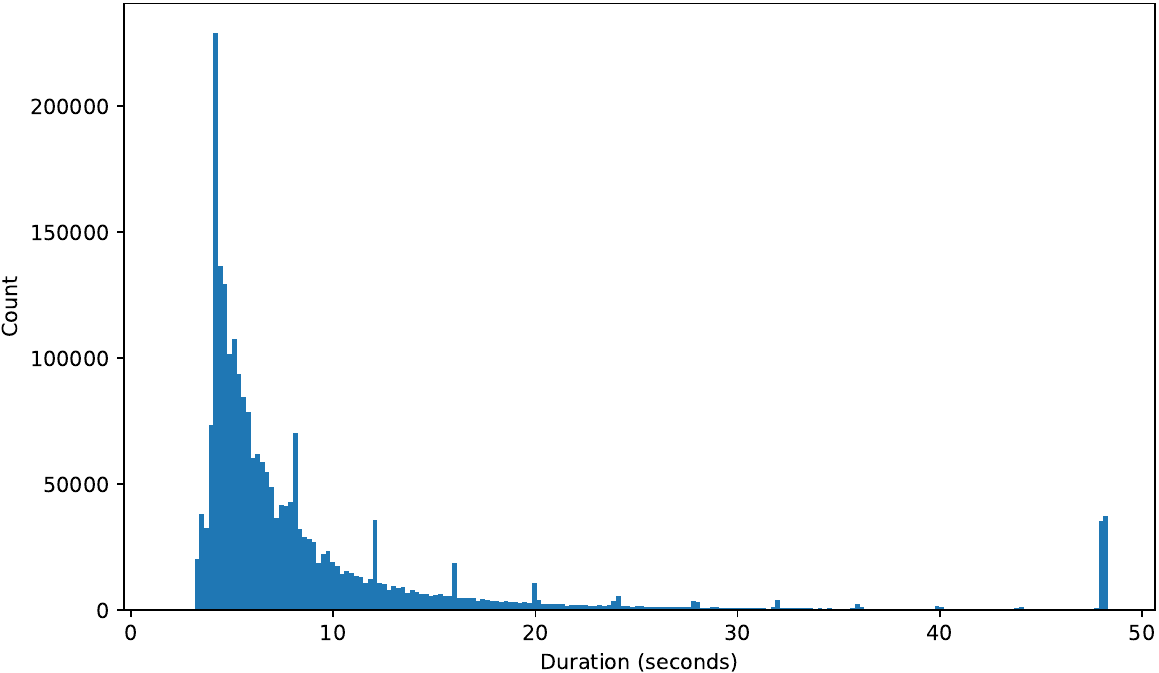}
        \vspace{-1mm}
        \label{fig:stat_duration}
    \end{subfigure}
    \hfill
    \begin{subfigure}{0.25\linewidth}
        \centering
        \includegraphics[width=0.99\textwidth]{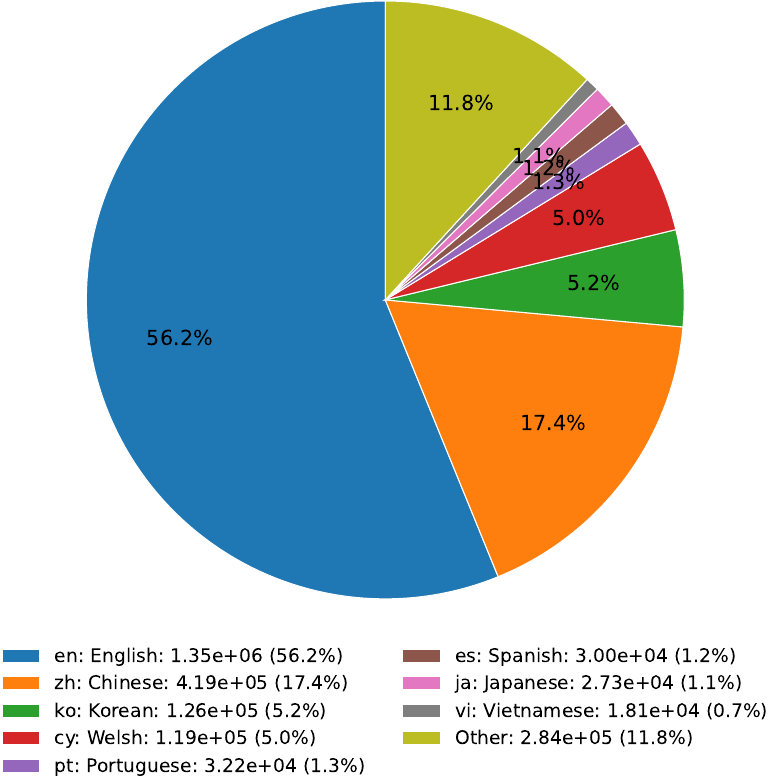}
        \vspace{-1mm}
        \label{fig:stat_language}
    \end{subfigure}
    \hfill
    \begin{subfigure}{0.22\linewidth}
        \centering
        \includegraphics[width=0.99\textwidth]{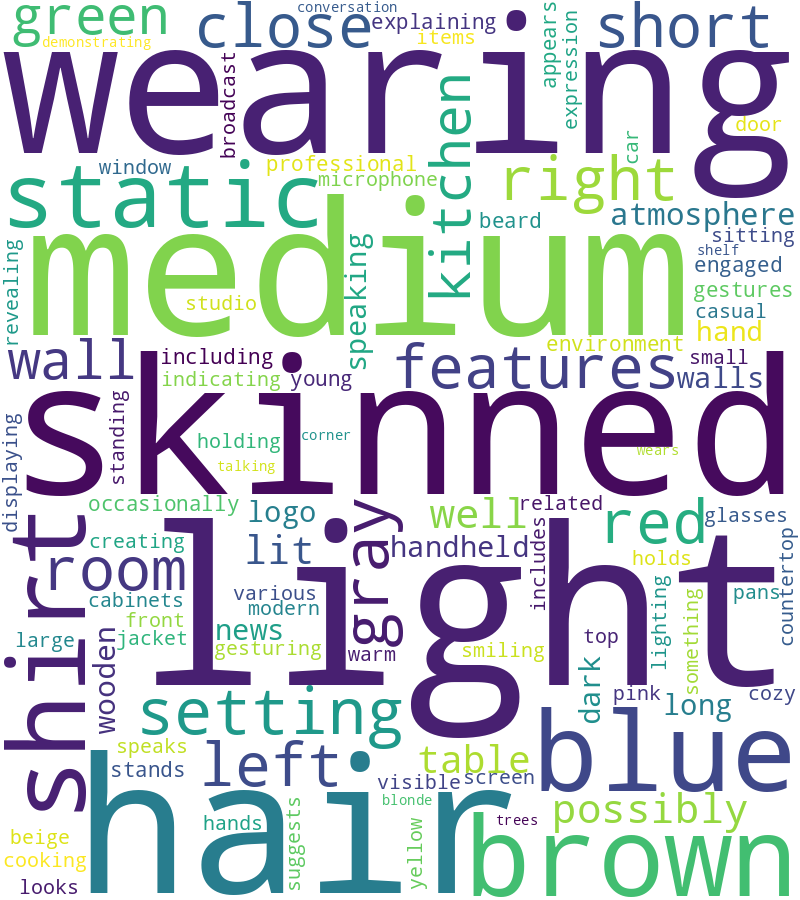}
        \vspace{-1mm}
        \label{fig:stat_wordcloud}
    \end{subfigure}
    \vspace{-1em}
    \caption{Dataset statistics and analysis: (left) distribution of clip durations; (middle) language distribution from automatic detection on speech tracks; (right) keyword word cloud summarizing caption content.}
    \label{fig:dataset_stats}
    \vspace{-1em}
\end{figure*}

\section{TalkVerse Dataset}
\label{sec:dataset}
    
TalkVerse comprises \textbf{2.3M} short clips (5--50s) totaling \emph{6.3k} hours, each with (i) 720p/1080p frame sequences, (ii) time-aligned speech audio, (iii) 2D pose skeletons, (iv) visual and audio captions, and (v) metadata for analysis and evaluation such as duration, language type, and quality scores. We curate primarily from \emph{OpenHumanVid}~\cite{li2024openhumanvid} and \emph{Panda-70M}~\cite{chen2024panda}. All clips are standardized, de-duplicated, and verified for audio--visual alignment prior to downstream processing. To reduce active-speaker ambiguity for audio-driven generation, we restrict clips to \emph{a single visible person} and talking/performing segments.

\subsection{Data Processing}
As shown in Fig.~\ref{fig:data_pipeline}, we construct a comprehensive data filtering pipeline to ensure the high visual quality and audio synchronization of our dataset.
\subsubsection{Video Preprocessing}
\noindent\textbf{Video Segmentation.}
We first apply codec standardization (H.264) and black-border removal. All videos are converted to 25 fps for the convenience of lip-sync filtering. Scene splitting is performed using PySceneDetect~\cite{Castellano_PySceneDetect} to segment videos into coherent shots, after which we retain segments of 5--50 seconds to balance semantic completeness and training efficiency. Notably, many internet videos may have scene boundaries that cannot be detected by PySceneDetect~\cite{Castellano_PySceneDetect} due to edited videos recorded from a static camera, where the background is continuous but the foreground may have sudden posture changes. This leads to unexpected human movements within a video generation window (e.g., 5s) in long-video generation when training an audio-driven video generation method upon it. For this type of scene boundary, we use AutoShot~\cite{zhu2023autoshot} to cut segments to ensure temporal smoothness. Empirically, we find it is effective for audio-driven video generation training to remove sudden posture changes in the generated results.

\noindent\textbf{Video Quality Filter.}
We perform a multi-faceted quality screening that considers luminance, blurriness, aesthetic and image quality~\cite{wu2023q}. Thresholds are selected to achieve high recall at intake and high precision post-filtering.

\subsubsection{Human-Centric Annotation}
\noindent\textbf{Textual Caption.}
To preserve strong text-conditioning for audio-injection training, we additionally produce \emph{structured} video captions covering background appearance, camera movement, subject appearance, and subject movement with a fine-tuned Qwen2.5-VL~\cite{bai2025qwen2} from seven evenly spaced frames from the clip, and an \emph{audio-style} caption with Qwen3-Omni~\cite{xu2025qwen3} from the entire audio track. The audio description emphasizes speaker-related acoustic attributes such as age, gender, prosody, emotion, speaking rate, as well as the ASR results. An audio-style caption can be beneficial for emotion consistency in long-video generation. Note that for audio-driven video generation we do not include ASR information in training, yet it could facilitate joint audio--video generation in the future. Additionally, we will leverage the OCR information from visual captions to remove videos with moving subtitles like broadcasts or TV shows. Videos with static text overlays are kept as long as they stay in the same position across the whole video.

\noindent\textbf{Skeleton Sequence.}
We extract human skeletons with DWPose~\cite{dwpose} to serve as pose control signals for human video generation. In addition, we use the detected person bounding box size and head/hand keypoint confidences to filter the data: we keep clips with a single person and persistent head keypoints, require the bounding box area $>\!0.2\times$ the whole image area, enforce head confidence $>\!0.9$, and, when hands are present, hand keypoint confidence $>\!0.8$. As a result, our dataset can contribute to pose-driven video generation with million-level video data.

\noindent\textbf{Speech Audio.}
We use SyncNet~\cite{syncnet} to check lip--audio synchrony and remove async videos, strengthening audio-conditioned training. Specifically, SyncNet learns a joint audio--visual embedding and outputs a scalar synchronization confidence and a temporal offset (in frames). We run it at scale on millions of clips and retain clips satisfying $|\text{offset}|\!\le\!3$ and $\text{confidence}\!>\!1.6$. We observed that even a small amount of out-of-sync data degrades lip-sync learning in audio-driven video generation, so we adopt these strict thresholds to minimize misalignment.
 
\begin{figure*}[t]
    \centering
    \includegraphics[width=0.99\textwidth]{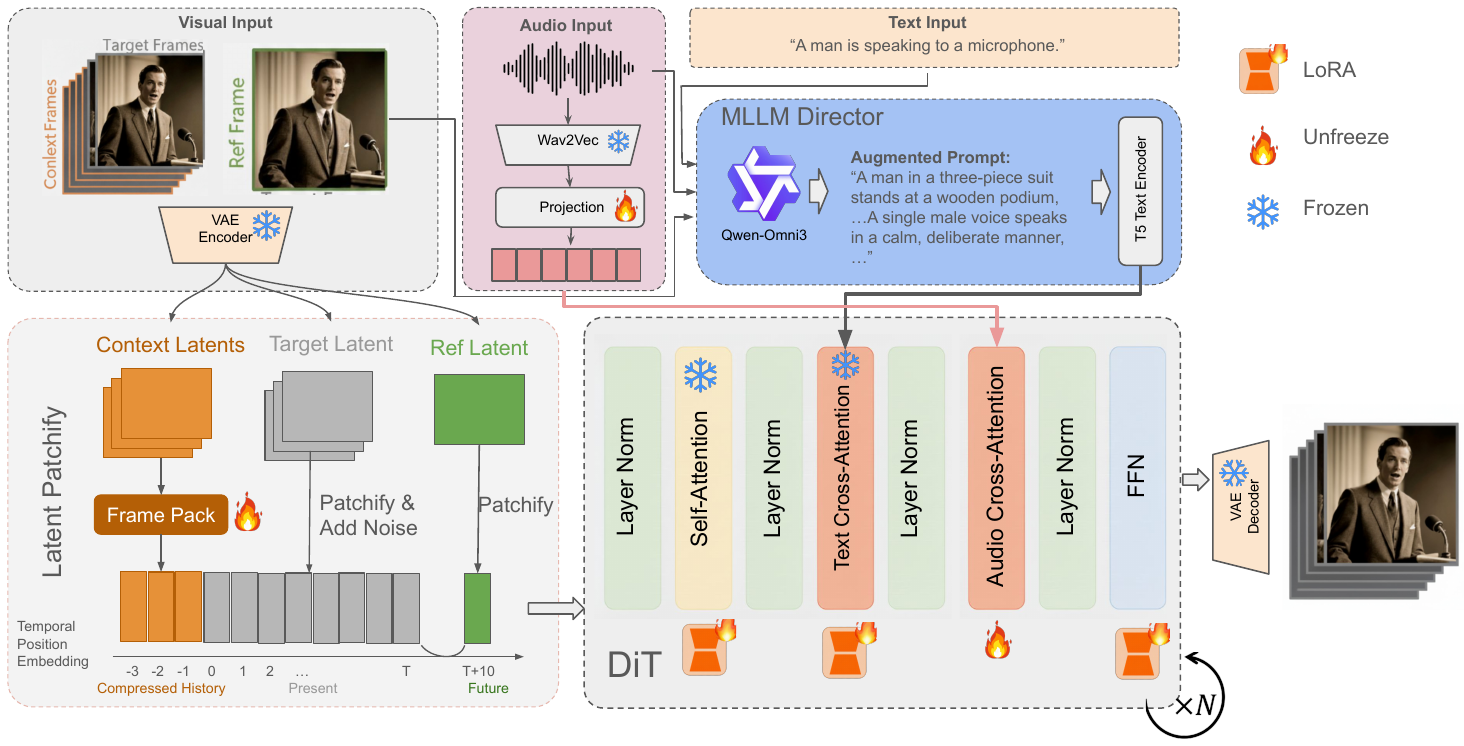}
        \vspace{-1em}
    \caption{TalkVerse‑5B model architecture. Wav2Vec features are projected and injected via sparse audio cross‑attention. FramePack compresses context; a reference‑image positional embedding prevents drift across sliding windows. An MLLM Director rewrites prompts conditioned on audio and reference frames. VAE uses (t,h,w) = (4,16,16) downsampling for a 4× token reduction versus typical settings.}
    \label{fig:model}
    \vspace{-1em}
    \end{figure*}
    
\section{Method}
\label{sec:baseline}
\subsection{Audio-Driven Video Generation}
Given a reference image, an audio input, and a text prompt describing the video content and audio style, we can generate a video synchronized with the audio while preserving the content in the reference image, as shown in Fig.~\ref{fig:model}. Our base model (Wan2.2-5B) is a Diffusion Transformer (DiT) pretrained for text/image-to-video generation, paired with a video VAE that downsamples by $(t,h,w)=(4,16,16)$, which has a larger compression ratio than widely used ones (e.g., Wan2.1-14B). As a result, this latent space is 4$\times$ faster than previous ones when generating the same video resolution. Concurrent with Wan-S2V~\citep{gao2025wan}, we find that using a set of \{motion-frame context, denoised video, reference image\} with proper positional embedding can generate smooth and coherent minute-long videos via a sliding window. The visual quality, subject consistency, and anti-drifting capability of this design are generally better than previous methods with first-frame image conditioning~\cite{kong2025let,gan2025omniavatar}. In training, the video duration only needs to be the length of motion-frame context plus denoised video, e.g., 3s + 4s in our case, much shorter than the minute-long objective.

\subsubsection{Audio Condition Injection}
We encode raw audio with Wav2Vec~\cite{baevski2020wav2vec2}, aggregate multi-layer Wav2Vec features via projection layers to fuse audio cues from different semantic levels, and compress them with causal 1D convolutions so that each latent frame receives a temporally aligned audio segment. Audio features are injected into DiT blocks by per-latent cross-attention from visual tokens to audio tokens, ensuring precise lip--audio synchronization. The audio cross-attention is applied to selected DiT blocks sparsely to reduce the computational overhead, and is trained in an end-to-end manner.

\subsubsection{Long-Video Generation}
\noindent{\bf Image Condition and Anti-drifting in Long Videos.} 
We find that the image condition is crucial for the visual appearance consistency in long videos. We adopt a Positional Embedding slightly longer than the last latent of denoised video (e.g., 10 latents) to guide the generation. This assignment makes the generation “chase” the reference image in the future rather than copy it at the current step, effectively decoupling the human posture in the reference image from the posture in the training video and avoiding copy-paste artifacts. As a result, we can share the same reference image across sliding windows at inference without inducing looped body movements; in contrast, first-frame conditioning tends to reset the starting latent of each window to exactly match the reference pose. Finally, this scheme also supports injecting different reference images for different windows to gradually change the background while keeping the subject's appearance consistent, yielding more cinematic long videos.

\noindent{\bf Context History Frames.} Following Wan-S2V~\citep{gao2025wan}, we pass a window of preceding motion frames as context and pack them with a FramePack module~\cite{zhangframe} to preserve smooth transition between each short video clip (e.g., 5 seconds). FramePack could effectively compress the motion frames and reduce the total number of tokens in the expensive 3D self-attention (temporal, width and height). The patchify embeddings are trained in a data-driven manner jointly with the audio cross-attention training. To make the set of \{motion-frame context, denoised video, reference image,\} consistent in positional embedding, we assign negative positional embedding to the motion frames to indicate that the context is the history information.

\begin{figure}[t]
    \centering
    \includegraphics[width=0.5\textwidth]{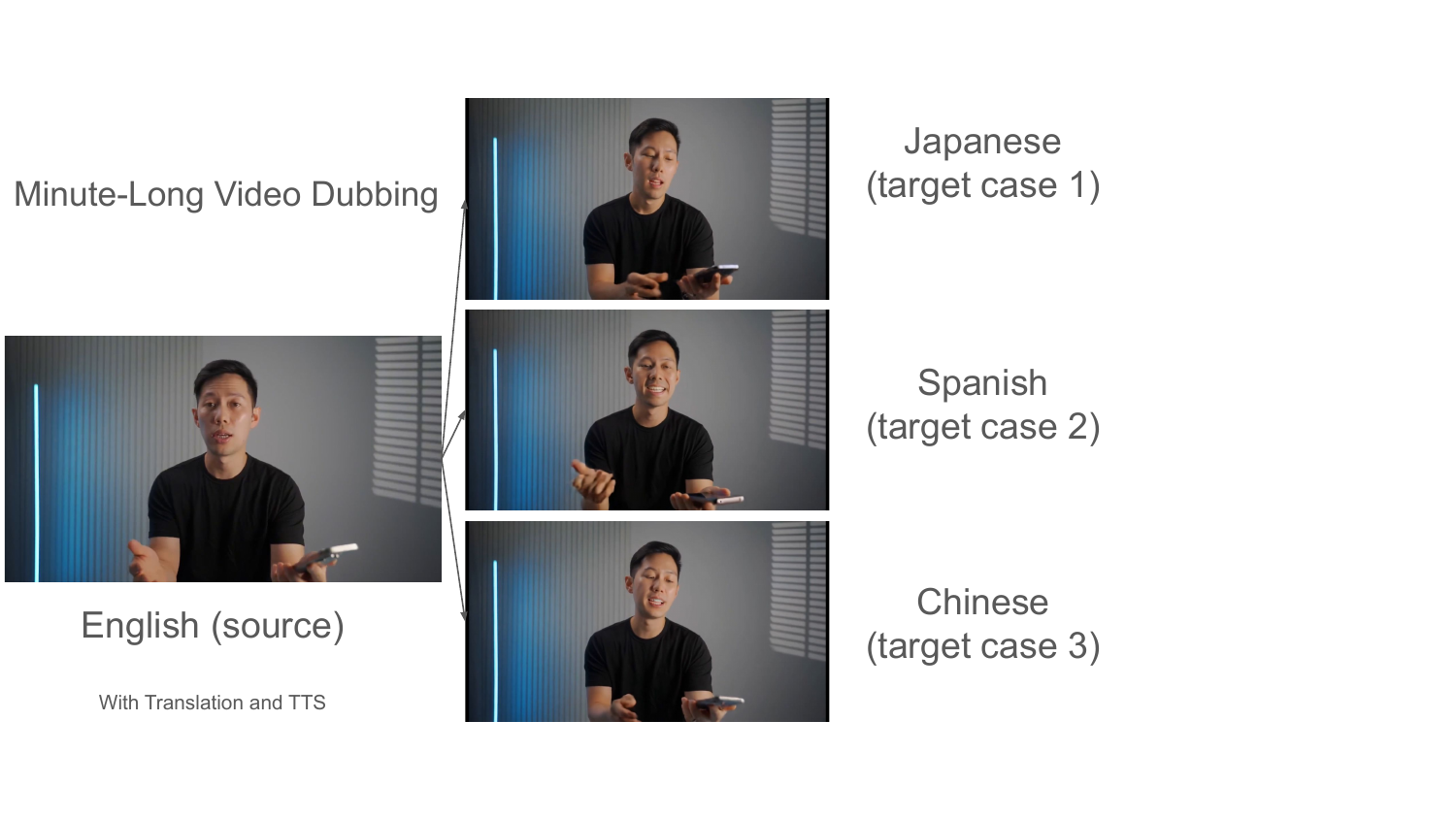}
        \vspace{-2em}
    \caption{Illustration of video dubbing using the pretrained audio-driven reference-image-to-video generation model with video editing tricks during inference.}
    \label{fig:dubbing}
    \vspace{-1em}
\end{figure}

\subsubsection{MLLM Director}
In minute-long video generation, there may be storylines in text prompts that guide the generation. Therefore, an MLLM Director that accepts user prompts and audio/image features to rewrite and produce a diverse and accurate video caption is crucial for the generation. In this stage, the rewritten video caption can also reduce the training--inference gap by matching the text prompt style in the training stage. To ground multimodal instructions into a coherent plan, we use Qwen3-Omni~\cite{xu2025qwen3} as an MLLM Director. It extracts emotion from the audio and scene/appearance descriptors from reference frames, then merges them with the user prompt to produce a compact storyline. The storyline prioritizes user intent, then audio cues, then image references, and enumerates character traits, background layout, actions, emotional shifts, visual style, and camera planning. We employ few-shot in-context templates to stabilize outputs. The finalized storyline is injected as text conditioning via cross-attention to guide generation consistently across time. As a result, the trained audio-driven DiT model can ensure lip synchronization, visual appearance consistency, and smooth window transitions in long videos, while the MLLM Director can guide the generation to be interesting and consistent with the user prompt and audio/emotion style.

\subsubsection{Training techniques.}
As Wan2.2-5B with a higher compression ratio VAE is more prone to generating artifacts during fine-tuning, we find that full parameter tuning is not suitable for the 5B model, which is different from the results from 14B models~\cite{gao2025wan}. We perform audio-driven fine-tuning via Low-Rank Adaptation (LoRA)~\cite{hu2021lora} for all DiT weights and full-parameter fine-tuning for audio projection, cross-attention, and FramePack-related weights. Specifically, for a pretrained weight matrix $W$, we introduce a low-rank residual $\Delta W$ and form the adapted weight $W' = W + \Delta W$, where $\Delta W$ is factorized as
$\Delta W = AB^\top,\quad A \in \mathbb{R}^{n \times d},\; B \in \mathbb{R}^{m \times d},\; d \ll \min(n,m)$.
By training only the small matrices $A$ and $B$ (and keeping $W$ frozen) in attention and MLP blocks inside DiT blocks, we greatly reduce memory and compute while preserving the base model's priors. This enables more stable adaptation to synchronized speech cues with minimal visual artifacts. To facilitate faster convergence of audio-conditioning injection, we initialize the audio cross-attention layers from the text cross-attention weights instead of random initialization. The major difficulty of model training is to add audio conditioning ability and smooth context frame transitions without damaging the text-conditioning ability. To prevent the new layers from disturbing the pretrained model's knowledge, we use a zero-initialized linear layer to connect the audio cross-attention layer to the main DiT blocks like ControlNet~\cite{DBLP:conf/iccv/ZhangRA23}. LoRA training is also beneficial for preserving the text-conditioning priors.

\begin{table}[t]
    \centering
    \caption{Quantitative comparisons on EMTD test set. $^\dagger$ our experimental results from Wan-S2V-14B official checkpoint.}
    \vspace{-1em}
    \label{tab:quantitative_comp}
    \setlength{\tabcolsep}{2pt}
    \resizebox{\columnwidth}{!}{
    \begin{tabular}{lcccccc}
      \toprule
      Method & FID$\downarrow$ & FVD$\downarrow$ & \small{Sync-C}$\uparrow$ & HKC$\uparrow$ & HKV$\uparrow$ & CSIM$\uparrow$ \\
      \midrule
      EchoMimicV2~\citep{DBLP:conf/cvpr/MengZLM25} & 33.42 & 217.71 & 4.44 & 0.425 & 0.150 & 0.519 \\
      EMO2~\citep{emo2} & 27.28 & {129.41} & 4.58 & 0.553 & {0.198} & {0.650} \\
      FantasyTalking~\citep{wang2025fantasytalking} & 22.60 & 178.12 & 3.00 & 0.281 & 0.087 & 0.626 \\
      Hunyuan-Avatar~\citep{chen2025hunyuanvideo} & 18.07 & 145.77 & 4.71 & 0.379 & 0.145 & 0.583 \\
      Wan-S2V-14B~\citep{gao2025wan} & 15.66 & 129.57 & 6.41$^\dagger$ & {0.435} & 0.142 & 0.677 \\
      \midrule
      TalkVerse-5B & 18.66 & 147.21 & 5.47 & {0.412} & 0.151 & 0.658 \\
      \bottomrule
    \end{tabular}
    }
    \vspace{-1em}
  \end{table}

\begin{figure*}[t]
    \centering
    \includegraphics[width=0.99\textwidth]{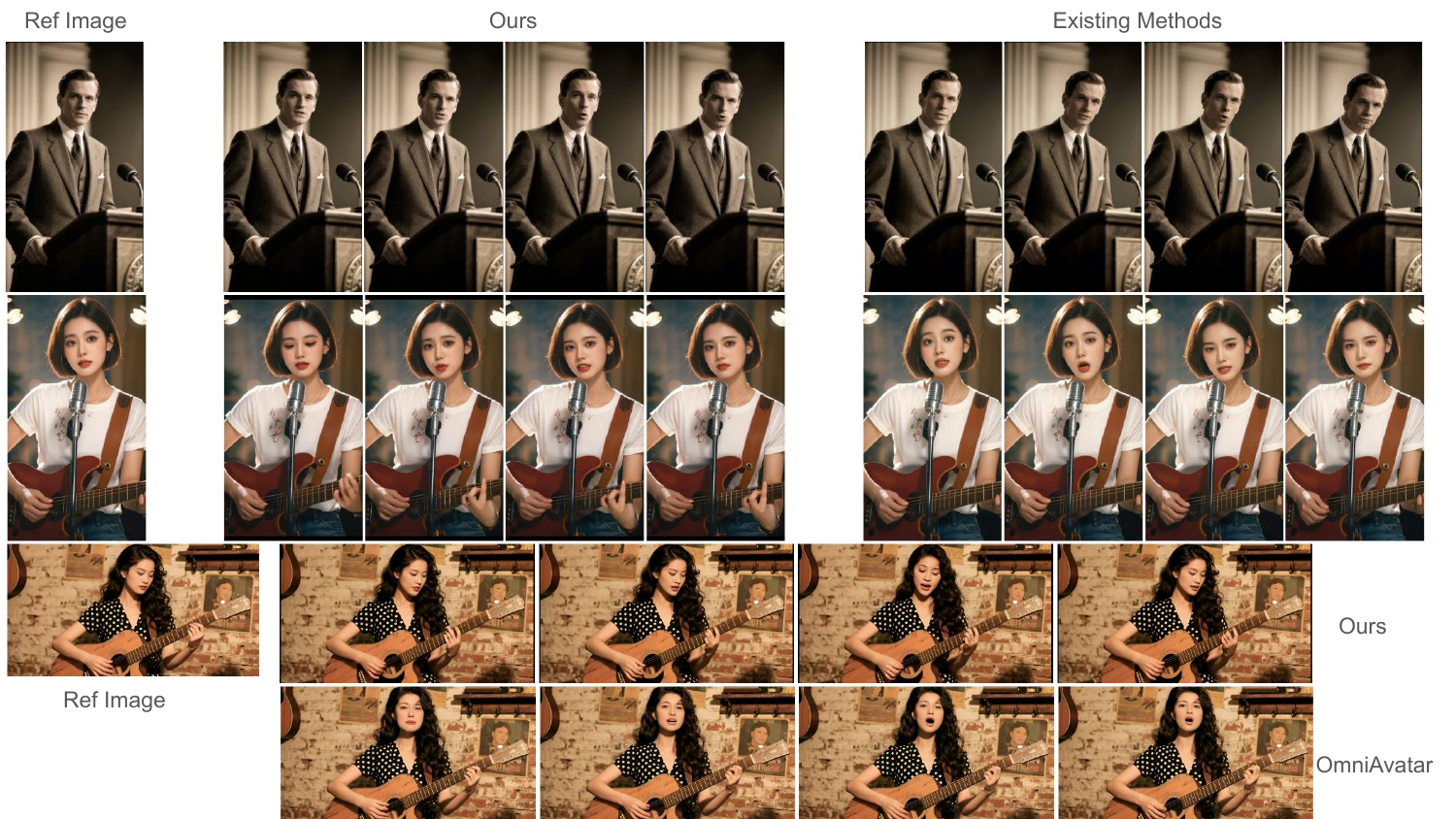}
        \vspace{-1em}
    \caption{Qualitative comparison with EchoMimicV3~\cite{meng2025echomimicv3} (case 1\&2) and OmniAvatar~\cite{gan2025omniavatar} (case 3), from top-down order.}
    \label{fig:sota}
    \vspace{-1em}
\end{figure*}

\subsection{Application to Video Dubbing}
We find that once an audio-driven image-to-video generation model is trained, it can generalize to audio-driven video-to-video generation in a zero-shot manner. As shown in Fig.~\ref{fig:dubbing}, given an input video, we adapt our generation pipeline to enable flexible control over content preservation and motion synthesis. Rather than initializing the diffusion process with pure Gaussian noise, we inject controlled noise into the encoded video latents at a specified noise level $\alpha \in [0, 1]$. Specifically, we first encode the input video into the latent space using our VAE encoder, then initialize the denoising process with:
\begin{equation}
\mathbf{z}_t = (1 - \alpha) \mathbf{z}_0 + \alpha \boldsymbol{\epsilon}, \quad \boldsymbol{\epsilon} \sim \mathcal{N}(0, \mathbf{I}),
\end{equation}
where $\mathbf{z}_0$ represents the clean video latents and $\mathbf{z}_t$ is the noisy initialization. The denoising process then proceeds only from timestep $t = \alpha T$ to $t = 0$, where $T$ is the total number of timesteps. 

We empirically set $\alpha = 0.95$ as the default value, as we observe that motion dynamics are predominantly synthesized during high-noise denoising steps in flow-based diffusion models. This high noise level allows the model to generate new audio-synchronized motion and expressions while the remaining low-noise steps refine details and preserve scene structure from the original video. The noise strength $\alpha$ provides intuitive control over the dubbing effect: lower values ($\alpha \to 0$) preserve more original video content, while higher values ($\alpha \to 1$) enable more flexible motion generation. Additionally, for temporal consistency across segments during multi-clip generation, we dynamically update the reference image by randomly sampling a frame from each video segment, ensuring the reference conditioning remains aligned with the current scene context throughout the entire video. 
\section{Experiments}
\subsection{Implementation Details}
We follow Wan2.2-5B and use resolution bins matching 720p area during training. Starting from a pretrained Wan2.2-5B, we train newly introduced parameters (i.e., audio projection, cross-attention, and FramePack-related weights) from scratch on TalkVerse while keeping DiT frozen, applying LoRA tuning to DiT. Audio layers are injected sparsely at DiT blocks $3k$ for $k\!=\!0,\dots,9$, plus the final block, leading to a total of 5.8B parameters. Training uses DDP with up to 73 context + 80 video frames due to GPU memory limits, batch size 1 per GPU on 64 GPUs. LoRA rank/alpha are 128. We use AdamW with lr 1e-5 (full fine-tuning parameters) and 1e-4 (LoRA). For training FramePack, we use mixed-length sampling: if a video $\geq$ 73+80 frames, we randomly sample a 73+80 window (first 73 as context, last 80 as video); otherwise, we zero the context latents and randomly sample an 80-frame video window. We sample the reference image temporally after the video window to be consistent with the positional embedding in our method. We use CFG=6.5 for both text and audio. Due to compute resource limits, we train for only one week ($\sim$1 epoch) and report these results in this section. We believe the performance could be further optimized with more computations.

\begin{figure*}[t]
    \centering
    \includegraphics[width=0.99\textwidth]{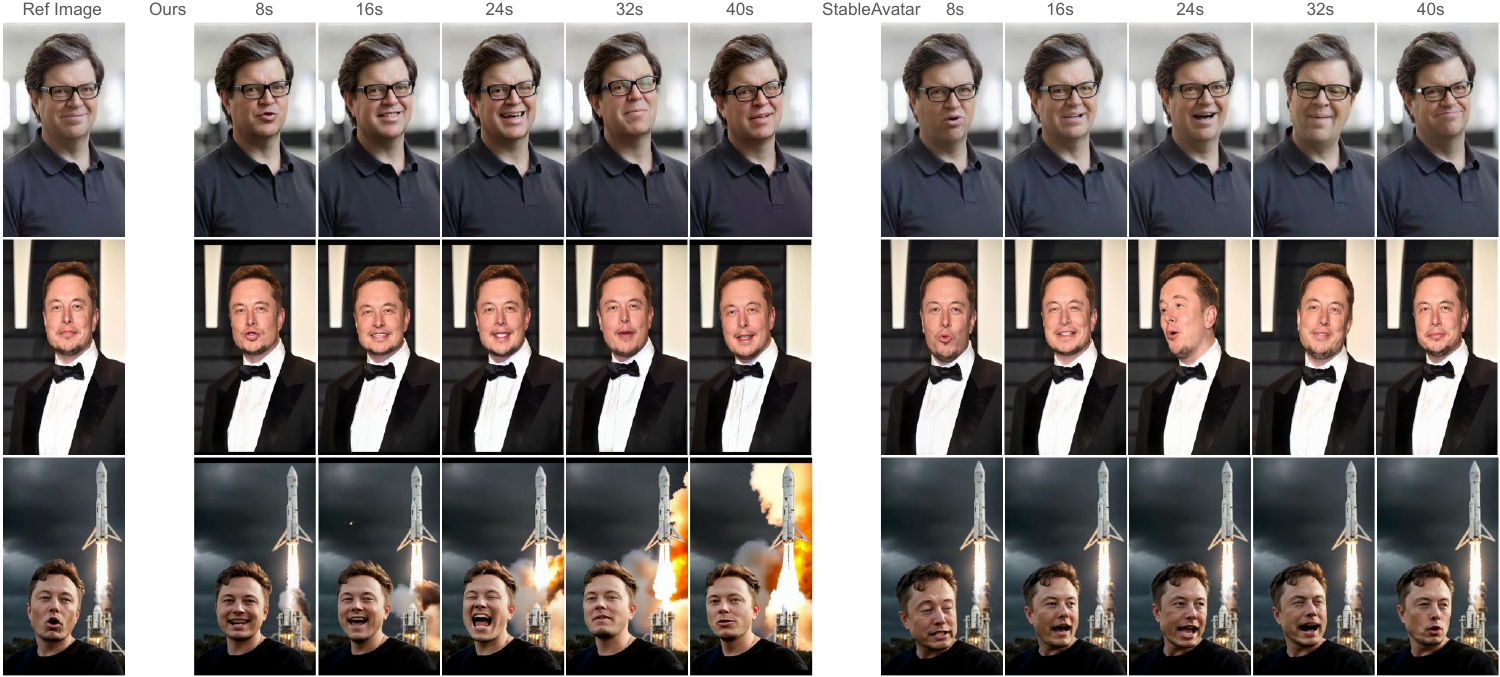}
        \vspace{-1em}
    \caption{Qualitative comparison on 40s long video generation (25fps) with StableAvatar~\cite{tu2025stableavatar}.}
    \label{fig:long_sota}
    \vspace{-1em}
\end{figure*}

\begin{figure}[t]
    \centering
    \includegraphics[width=0.35\textwidth]{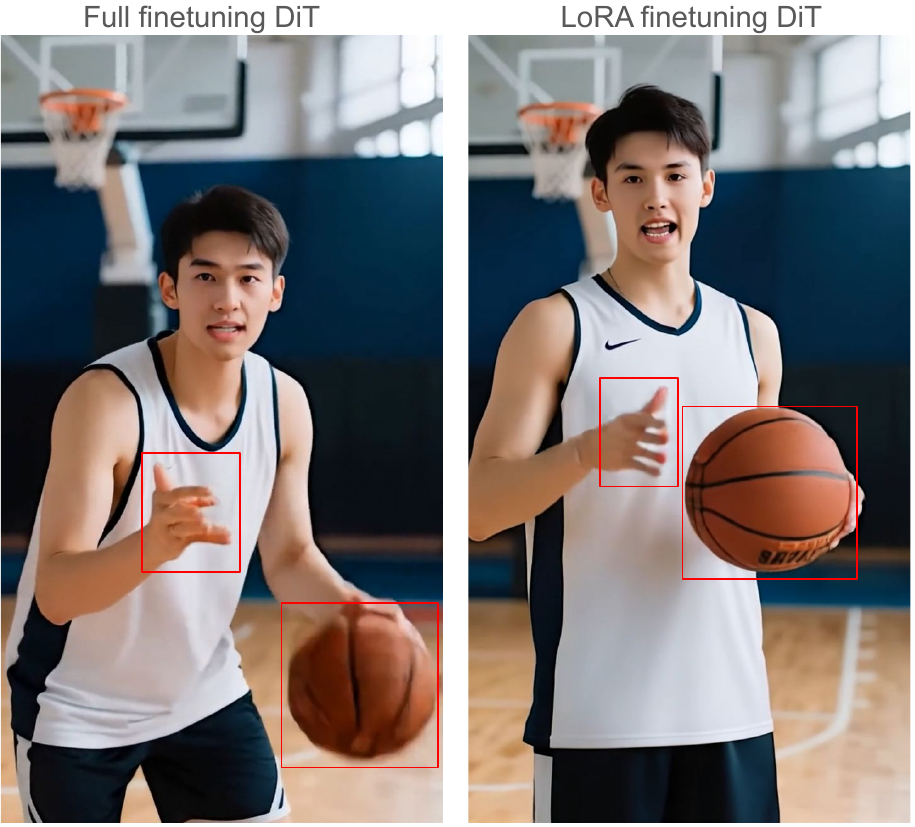}
        \vspace{-1em}
    \caption{Qualitative comparison on LoRA training and full finetuning on DiT weights.}
    \label{fig:lora_training}
    \vspace{-1em}
\end{figure}

\noindent{\bf Metrics.}
As most open-source benchmarks focus on talking heads, we evaluate on the EMTD dataset~\cite{DBLP:conf/cvpr/MengZLM25} and compare with EchoMimicV2~\cite{DBLP:conf/cvpr/MengZLM25}, EMO2~\cite{emo2}, FantasyTalking~\cite{wang2025fantasytalking}, and Hunyuan-Avatar~\cite{chen2025hunyuanvideo} following Wan-S2V-14B~\cite{gao2025wan}. We report FID~\cite{heusel2017gans}, FVD~\cite{unterthiner2019fvd}, lip-sync Sync-C~\cite{syncnet}, ArcFace~\cite{deng2019arcface} similarity (CSIM), and hand keypoint confidence (HKC) and hand keypoint variance (HKV). We remove PSNR and SSIM due to the fact that they are ineffective with audio condition.

\subsection{Main Results}
\noindent{\bf Lip-sync and Identity Consistency.} Our 5B model's performance is comparable to SOTA methods on FID/FVD/lip-sync/hand quality/identity consistency. Compared to Wan-S2V-14B and FantasyTalking based on Wan2.1-14B model, this performance is achieved with roughly $11.2\times$ lower inference compute, stemming from a $4\times$ reduction in latent tokens due to our higher VAE compression and an additional $\approx 2.8\times$ speedup from using 5B vs 14B parameters (14/5). Hunyuan-Avatar also uses a large 13B HunyuanVideo video model, which is significantly slower than ours. In Fig~\ref{fig:sota}, we show qualitative comparisons with EchoMimicV3~\cite{meng2025echomimicv3} and OmniAvatar~\cite{gan2025omniavatar}. Our model can generate more natural body movements and better subject consistency than open-sourced academic methods.

\noindent{\bf Long Video Generation.} We also evaluate our model on 40s long video generation at 25fps and compare with StableAvatar~\cite{tu2025stableavatar} in Fig.~\ref{fig:long_sota}. We can see that our model can achieve on-par long-video appearance preservation ability with StableAvatar after 1000 frames, which could cover most use cases in applications.

\subsection{Ablations}

\noindent{\bf Lora Training.} 
We compare LoRA training and full fine-tuning on DiT weights in Fig.~\ref{fig:lora_training}. We can see that LoRA training can preserve better visual quality than full fine-tuning on DiT weights, especially for articulated hands and objects. Our observation is that small pretrained DiT models (e.g., 5B or 1.3B) could suffer more from visual artifacts during fine-tuning than the powerful 14B model. However, the heavy 14B model could be prohibitively slow for inference in applications and impede broad participation in the community. We argue that the 5B model is a good compromise between performance and inference speed, with a suitable training strategy. Furthermore, it could have the potential to be real-time with proper distillation and engineering optimization.

\section{Conclusion}
We introduced TalkVerse, a large-scale, audio-synchronized video corpus for single-person human video generation conditioned on audio or pose. The dataset comprises 2.3M clips (6.3k hours) at 720p/1080p, each with time-aligned speech audio, DWPose skeleton sequences, and video/audio captions. On top of this data, we train a Wan2.2-5B DiT, yielding a substantially faster latent space than widely used settings. Despite its small size, the model attains perceptual quality and lip synchronization comparable to larger 14B systems with 10$\times$ lower inference cost, enabled by audio condition injection from aggregated Wav2Vec features via sparse cross-attention, FramePack-based history compression for smooth window transitions, and a reference-image positional embedding strategy that sustains minute-long generation with less drift. An MLLM Director (Qwen3-Omni) rewrites user prompts by grounding audio and image cues to produce coherent, style-consistent storylines. Our model can generalize zero-shot to video dubbing via controlled noise injection.

\noindent{\bf Limitations and future work.}
5B models may underperform on complex and extreme hand articulation and full-body motions. Although this dataset could support other tasks like pose-driven video generation and audio-video joint generation, this paper does not include these experiments due to compute resource limits. We will adapt our dataset to these tasks in future work.

\clearpage
\setcounter{page}{1}
\maketitlesupplementary

\section{More Details about Experiments}
\label{sec:supp}

\subsection{Training Details}

We finetune the Wan S2V 5B model and newly added audio cross-attention using AdamW ($10^{-5}$ learning rate) with selective unfreezing of the time embedding, projection, head, and optional transformer blocks. LoRA parameters in DiT are scaled by $10\times$ to be $10^{-4}$. Training uses a batch size of 1 per GPU with gradient accumulation to be total batch size as 256 and gradient clipping (norm 1.0). To perform CFG, we apply 10\% dropout to text, image, and audio conditions. When framepack module is enabled, we jointly encodes motion and target frames to VAE space for training to maintain temporal continuity in long-video generation. Otherwise, the motion context frames in VAE space as all-zero tensors.

\subsection{Spatial Loss Weighting}
We use a ROI loss up weights for reconstruction error in face and body regions by employing DWPose~\cite{dwpose} to estimate body and face bounding boxes and keypoints which are normalized and rasterized into binary masks aligned with the resolution in VAE space, and then construct a spatial loss weighting over the diffusion denoising objective. Specifically, let $L_{\mathrm{full}}$, $L_{\mathrm{body}}$, and $L_{\mathrm{face}}$ denote the per-pixel losses on the full frame, the body bounding box, and the face bounding box, respectively. The final training loss is
\begin{equation}
    L = \frac{1 \cdot L_{\mathrm{full}} + 1 \cdot L_{\mathrm{body}} + 1 \cdot L_{\mathrm{face}}}{Z},
\end{equation}
where $Z$ is a normalization factor (e.g., the sum of the weights), so that regions inside the body and face boxes receive proportionally higher importance than the background.

\subsection{Data Preprocessing Pipeline}

We construct a video--audio--text dataset by unifying metadata from multiple sources, ensuring all clips contain valid 25\,fps video and audio streams. To control resolution, we adopt a resolution-bin policy (e.g., 720p) where each clip is resized and center-cropped to the closest target bucket. We sample a continuous sequence of frames for denoising, optionally prepended by motion context frames, and extract a conditional reference image from a nearby window.

Audio is resampled to 16\,kHz and strictly aligned to the video segment, padded if necessary. Finally, we load cached text embeddings (pre-extracted by T5 text encoder) and a null embedding for classifier-free guidance, yielding a comprehensive set of synchronized inputs for training.

\subsection{Inference Process}

Our inference pipeline supports both image-to-video generation and video-to-video ``dubbing'' across multiple resolutions. We encode the text prompt (T5), audio (wav2vec2), and reference image (VAE) to condition the diffusion model. For dubbing, instead of initializing from pure noise, we encode the input video into latents, add noise to a target timestep $t_{\text{dub}}$, and denoise from that state to preserve coarse structure while synchronizing lip motion. The latent sequence is iteratively denoised using a flow-matching solver (e.g., UniPC, 50 steps) and decoded by the VAE. For CFG scale, we find that 3 to 6 will leads to good results, which is a trade-off between (1) visual quality and motion strength, and (2) lip-sync quality. The larger CFG weights, the better lip-sync quality but worse visual quality and motion strength.

\subsection{Hyper-parameter Settings for Wan S2V 5B}
We summarize the main configuration used in our experiments in Table~\ref{tab:wan_s2v_5b_hparams}.

\begin{table}[t]
    \centering
    \caption{Hyper-parameters for our 5B model.}
    \label{tab:wan_s2v_5b_hparams}
    \resizebox{1.0\linewidth}{!}{%
    \begin{tabular}{ll}
        \toprule
        Hyper-parameter & Value \\
        \midrule
        Language Encoder & T5-XXL \\
        VAE & Wan2.2-VAE \\
        VAE stride & $(4, 16, 16)$ \\
        Audio encoder & wav2vec2-large-xlsr-53-english \\
        Patch size & $(1, 2, 2)$ \\
        Model dimension & $3072$ \\
        FFN dimension & $14336$ \\
        Frequency embedding dimension & $256$ \\
        Attention heads & $24$ \\
        Transformer layers & $30$ \\
        QK norm / cross-attn norm & True / True \\
        Audio dimension & $1024$ \\
        Audio injection layers & $\{0, 3, 6, 9, 12, 15, 18, 21, 24, 27, 29\}$ \\
        Motion frames & $73$ \\
        Latent dimension & $48$ \\
        $v$-scale & $1.0$ \\
        Sampling frame rate & $25$ fps \\
        Sampling frame count & $120$ \\
        Diffusion shifting & $5$ \\
        Default sampling steps & $50$ \\
        \bottomrule
    \end{tabular}
    }
    \vspace{-1em}
\end{table}
\clearpage
{
    \small
    \bibliographystyle{ieeenat_fullname}
    \bibliography{main}
}

\end{document}